%% file: main.tex
\documentclass[10pt,twocolumn,letterpaper]{article}

\usepackage{cvpr}              

\input{preamble}

\usepackage{booktabs} 
\usepackage{multirow} 
\usepackage{algorithm}
\usepackage{algpseudocode}
\usepackage{adjustbox}
\usepackage[accsupp]{axessibility} 
\usepackage{xcolor}
\usepackage{soul}
\usepackage{threeparttable}
\usepackage{stfloats}
\usepackage{multirow}
\usepackage{array}
\newcolumntype{P}[1]{>{\centering\arraybackslash}p{#1}}
\usepackage{longtable}
\definecolor{cvprblue}{rgb}{0.21,0.49,0.74}
\definecolor{lightred}{HTML}{ff9999} 
\definecolor{myorange}{HTML}{f3a977}
\definecolor{mypink}{HTML}{EC008C}
\definecolor{blueone}{HTML}{d2def1}
\definecolor{yellowone}{HTML}{fff0c2}
\definecolor{greenone}{HTML}{dcecd2}

\definecolor{lightgreen}{HTML}{90EE90} 
\usepackage[pagebackref,breaklinks,colorlinks,citecolor=cvprblue]{hyperref}

\def\paperID{8317} 
\def\confName{CVPR}
\def\confYear{2024}

\title{Decomposing Disease Descriptions for Enhanced Pathology Detection: A Multi-Aspect Vision-Language Pre-training Framework}
\author{
Vu Minh Hieu Phan\textsuperscript{1}, Yutong Xie\textsuperscript{1}, Yuankai Qi\textsuperscript{2}, Lingqiao Liu\textsuperscript{1}, Liyang Liu\textsuperscript{1}, Bowen Zhang\textsuperscript{1},  \\Zhibin Liao\textsuperscript{1}, Qi Wu\textsuperscript{1}, Minh-Son To\textsuperscript{3},  Johan W. Verjans\textsuperscript{1} \\
\small \textsuperscript{1}Australian Institute for Machine Learning, The University of Adelaide; \small \textsuperscript{2}Macquarie University; \small \textsuperscript{3}Flinders University 
\\\small\textsuperscript{1}{\tt
\{vuminhhieu.phan,yutong.xie,lingqiao.liu,akide.liu,b.zhang,zhibin.liao,qi.wu01,}
\\\small{\tt
johan.verjans\}@adelaide.edu.au},
\small\textsuperscript{2}{\tt
yuankai.qi@mq.edu.au}, \small\textsuperscript{3}{\tt
minhson.to@flinders.edu.au}
}

\begin{document}
\maketitle
\input{sec/0_abstract}
\input{sec/1_intro}
\input{sec/2_related}
\input{sec/3_method}

\input{sec/4_result}
\input{sec/5_conclusion}
{
    \small
    \bibliographystyle{ieeenat_fullname}
    \bibliography{main}
}


\end{document}

%% file: preamble.tex
\usepackage[dvipsnames]{xcolor}


%% file: sec/0_abstract.tex
\begin{abstract}

Medical vision language pre-training (VLP) has emerged as a frontier of research, enabling zero-shot pathological recognition by comparing the query image with the textual descriptions for each disease. Due to the complex semantics of biomedical texts, current methods struggle to align medical images with key pathological findings in unstructured reports. This leads to the misalignment with the target disease's textual representation. In this paper, we introduce a novel VLP framework designed to dissect disease descriptions into their fundamental aspects, leveraging prior knowledge about the visual manifestations of pathologies. This is achieved by consulting a large language model and medical experts. Integrating a Transformer module, our approach aligns an input image with the diverse elements of a disease, generating aspect-centric image representations. By consolidating the matches from each aspect, we improve the compatibility between an image and its associated disease. Additionally, capitalizing on the aspect-oriented representations, we present a dual-head Transformer tailored to process known and unknown diseases, optimizing the comprehensive detection efficacy. Conducting experiments on seven downstream datasets, ours improves the accuracy of recent methods by up to 8.56\% and 17.26\% for seen and unseen categories, respectively. Our code is released at \href{https://github.com/HieuPhan33/MAVL}{https://github.com/HieuPhan33/MAVL}.

\end{abstract}

%% file: sec/1_intro.tex
\section{Introduction}
\label{sec:intro}
The advent of vision-language pre-training (VLP) methods, notably CLIP~\cite{radford2021learning}, has yielded impressive 
zero-shot and low-shot fine-tuning in object recognition~\cite{radford2021learning,yao2021filip,li2022blip,li2023blip}. Pre-training on 400M image-text pairs, VLP models~\cite{radford2021learning,yu2022coca,li2022blip,radenovic2023filtering} learn a strong semantic mapping between image and text spaces.
During inference, they enable zero-shot recognition by computing the similarity between the query image and the text representation of each category. 
In the medical field, recent methods~\cite{tiu2022expert,wu2023medklip,bannur2023learning,wang2022multi} adopt CLIP and align images with their corresponding reports to enable zero-shot disease recognition, \ie, classifying without further fine-tuning. As such, this strategy reduces the reliance on costly medical data annotation.
Adding supervisory signals can enhance discriminative features, exemplified by a concurrent work~\cite{wu2023medklip} that extracts disease terms from reports and directly matches with the disease's definitions.

\begin{figure}[!t]
    \centering
    \includegraphics[width=\linewidth]{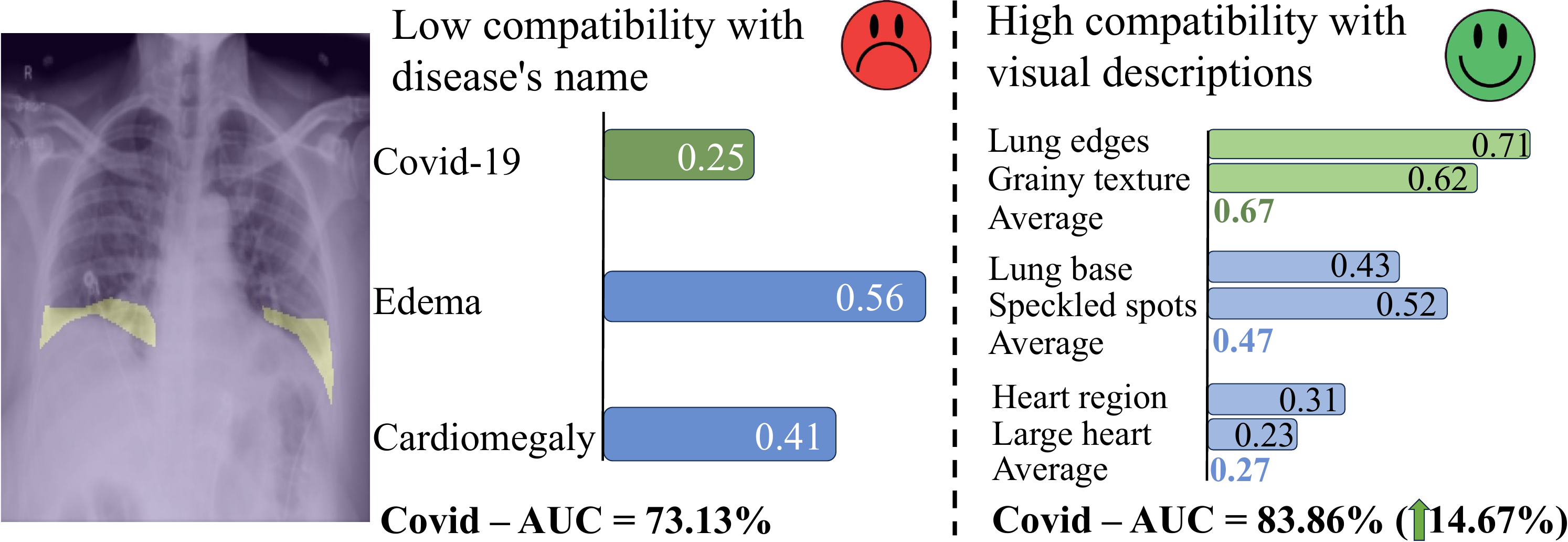}
    \caption{\textbf{Predictions of CheXzero~\cite{tiu2022expert} (Left), a strong CLIP-like model, and our multi-aspect matching model (Right).}  \texttt{Edema} and \texttt{Covid-19} belong to the domain of lung diseases, while cardiomegaly is a heart disease. CheXzero~\cite{tiu2022expert} misaligns the image feature with the target \texttt{Covid-19}, while it over-aligns with \texttt{edema}. We leverage the medical knowledge base to decompose disease terms into distinct visual components, enhancing the image alignment with the representations of the target disease.}
    \label{fig:intro}
    \vspace{-0.6em}
\end{figure}
\begin{figure}
    \centering
    \includegraphics[width=\linewidth]{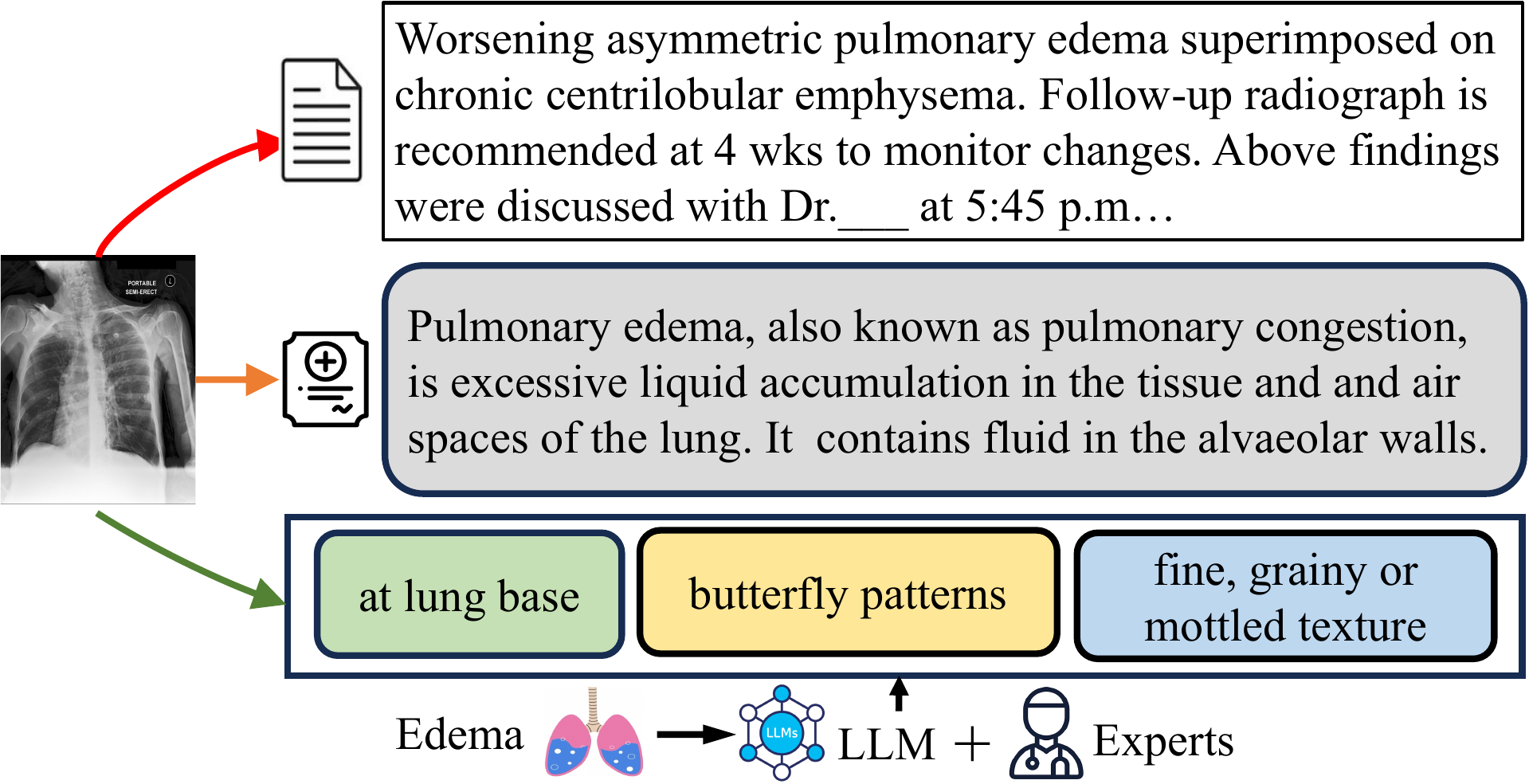}
    \caption{Illustrations of the three VLP paradigms: image-report matching~\cite{bannur2023learning,boecking2022making,tiu2022expert} (\sethlcolor{lightred}\hl{red} arrow), image-disease definition matching~\cite{wu2023medklip} (\sethlcolor{myorange}\hl{orange} arrow), and our proposed fine-grained image-aspect matching (\sethlcolor{lightgreen}\hl{green} arrow).}
    \label{fig:illustrate}
    \vspace{-0.6em}
\end{figure}

However, current VLP methods~\cite{tiu2022expert,wu2023medklip,bannur2023learning,wang2022multi} often produce low compatibility scores when aligning with the target disease's name, particularly for novel categories.
Fig.~\ref{fig:intro} (Left) shows the misalignment between an image showing previously unseen disease, \texttt{Covid-19}, and its target disease name.
This issue arises from various inherent challenges of VLP. \textit{First}, medical imaging analysis is inherently fine-grained, which requires distinguishing between diseases that are visually similar, 
such as differentiating  \texttt{Covid-19} from other lung diagnoses like \texttt{edema}. \textit{Secondly}, biomedical texts contain complex clinical terminologies, and there is an imbalance between important entities (\ie, medical findings) and less relevant texts~\cite{boecking2022making}. Fig.~\ref{fig:illustrate} shows sampled biomedical text inputs of current report matching~\cite{bannur2023learning,boecking2022making,tiu2022expert} and disease definition matching methods~\cite{wu2023medklip}.
Constrained by the scarcity of medical image reports~\cite{wu2023medklip}, 
data-intensive models like CLIP struggle to align \textit{fine-grained} visual features with \textit{key} disease textual representations. \textit{Thirdly}, current approaches represent text inputs as raw reports~\cite{tiu2022expert,boecking2022making} or disease descriptions~\cite{wu2023medklip},  
failing to capture generalized representations for novel medical findings. Consequently, the misalignment issue is exacerbated for diseases unmentioned in the pre-training dataset. 


To resolve the image-text misalignment for both seen and unseen categories, 
 this paper proposes to introduce a novel multi-aspect vision-language matching (MAVL) framework. We dissect disease descriptions into elemental aspects (\eg, opacity level, shape) and harness pre-established knowledge about the diseases' visual profiles. 
We present a semi-automatic pipeline to extract descriptions of visual aspects of 75 medical findings using a large language model (LLM), \textcolor{mypink}{GPT-4}, and meticulously curate by consulting two medical experts. Our solution is motivated by two observations. First, structuring the text inputs into fine-grained visual aspects enhances alignments with the target diseases. Second, new diseases can be described by elemental visual aspects of base diseases, e.g., Covid-19 has a texture of edema (grainy texture) and typically is located in a similar position as pneumothorax (lung edges). Thus, multi-aspect decomposition improves the recognition of unseen diseases by linking visual appearances of any new diseases with the base visual knowledge. From Fig.~\ref{fig:intro} (Right), our model effectively aligns the image's features with the target \texttt{Covid-19} by leveraging the common aspects and extrapolating visual descriptions of seen lung diseases (e.g., edema) to those of a novel category. 

Furthermore, we empirically observe that contrastive learning is non-optimal to capture discriminative features for differentiating fine-grained seen diseases. Capitalizing on the aspect representations, our \textit{secondary contribution} is to introduce a dual-head module dedicated to maximize the detection accuracy of seen and novel diseases. Each head is trained via a direct supervised loss and a contrastive loss. The supervised head maximizes the discriminative ability on base categories, while the contrastive head achieves performant zero-shot recognition of novel categories. 

Our main contributions in this work are:
\begin{itemize}
    \item A novel multi-aspect vision-language pre-training (MAVL) framework to improve the alignment between an image and textual representations of diseases, especially for unseen categories.
    To the best of our knowledge, this is the first study that exploits prior knowledge about pathological visual profiles 
    to optimize fine-grained disease recognition.
    \item A dual-head Transformer that is trained via supervised loss and contrastive loss. Simple yet effective, we show that decoupling the learning signals is effective to maximize zero-shot recognition of unseen diseases, while protecting the discriminatory power toward base categories.
    \item State-of-the-art performance on zero-shot classification and grounding on both seen and novel disease categories, especially on rare diseases in a long-tailed distribution, across \textbf{seven} downstream datasets. Through fine-tuning, our model significantly surpasses previous methods, establishing a strong foundational model for both zero-shot and low-shot learning.
\end{itemize}





%% file: sec/2_related.tex
\section{Related Works}
\label{sec:related}
\textbf{General vision-language pre-training.} Self-supervised VLP generates robust visual representations for various downstream tasks by exploiting joint vision-language supervision~\cite{radford2021learning,li2022blip,li2023blip,yao2021filip,radenovic2023filtering}. Current methods can be divided into two categories: a dual-stream ~\cite{radford2021learning,li2021align,jia2021scaling,radenovic2023filtering} with two encoders for each modality, or a single-stream~\cite{su2019vl,gan2020large,li2021unimo,huang2021seeing,bao2022vlmo} model, which enables deep V+L fusions. The supervised head in our network uses Transformer-based fusion layers to extract discriminative features.

\noindent\textbf{Medical vision-language pre-training.} Most medical VLP methods~\cite{bannur2023learning,boecking2022making,liu2023improving,wang2022medclip,zhang2022contrastive,huang2021gloria,wang2022multi} adopt the two-stream approach, which applies image-report contrastive learning without using a deep fusion module. 
ConVIRT~\cite{zhang2022contrastive} proposes a bidrectional contrastive loss for image-report matching. GLoRIA~\cite{huang2021gloria} and LoVT~\cite{muller2022joint} captures fine-grained alignments by optimizing the local contrastive loss between image regions and sentence-level tokens. Improving the biomedical text modelling, BioViL~\cite{boecking2022making} adds clinical vocabulary and augments report-specific data.
Subsequently, BioViL-T~\cite{bannur2023learning} extends the prior work by exploiting self-supervisory signals in a series of temporal images and reports. MGCA~\cite{wang2022multi} conducts a pseudo-entity level alignment by assuming each disease prototype can be represented as a cluster in the feature space. 
A concurrent work, MedKLIP~\cite{wu2023medklip}, proposes to extract disease entities from reports and performs matching with the diseases' description. Yet, by matching reports or diseases' clinical definitions with complex terminologies, existing methods tend to misalign the image with the target disease's representations. Given the scarcity of medical data pairs and complex biomedical texts, \textit{how to improve the image compatibility with the pathological text representation} is an open-ended question in VLP. 

\noindent\textbf{Textual prompt engineering.} Several techniques investigate task-specific prompt engineering to adapt VLP for image classification~\cite{radford2021learning,shu2022test}, object detection~\cite{shu2022test}, and visual question answering~\cite{wang2022ofa}. In medical VLP, GloRIA~\cite{huang2021gloria} generates a set clinical-specific prompts describing possible sub-types, severities, and locations for each disease class. CheXzero~\cite{tiu2022expert} constructs a binary prompt to obtain a binary classification of a disease. MedKLIP~\cite{wu2023medklip} obtains clinical descriptions of medical findings from UMLS knowledge base~\cite{bodenreider2004unified}. Current medical VLP works use descriptions with domain-specific terminologies. In contrast, our proposed method leverages pathological's visual descriptions, guiding the model to effectively detect diseases in images. 

\noindent\textbf{Visual description for explainable VL models.} Language concept bottleneck models (CBM)~\cite{yang2023language,oikarinen2023label,menon2023visual,yuksekgonul2023post,yan2023learning,Yun2022DoVP} are an emerging field in explainable AI, which leverage visual descriptions of an object category to interpret VL model's decisions. Yet, they focus on \textit{explaining} the pre-trained model in the general domain, while our framework leverages visual descriptions for \textit{pre-training} medical models. Furthermore, extracting visual descriptions of diseases are non-trivial tasks compared to objects in the natural world. Ours is complementary to CBM studies, which potentially enables explainability by providing scores for each disease's appearance when classifying their presence.

%% file: sec/3_method.tex
\section{Proposed Method}
\label{sec:method}

In this section, we first describe the VLP's reformulation, which enables supervised training (Sec.~\ref{sec:problem}). Subsequently, we present our multi-aspect vision-language matching (MAVL) framework, which decomposes disease entities into a set of visual aspects leveraging the knowledge of medical experts and LLMs (Sec.~\ref{sec:aspect}). 
And thirdly, capitalizing on the aspect representations, Sec.~\ref{sec:transformer} introduces our dual-head Transformer model, maximizing the complementary detection of both unseen and seen pathologies. 

\subsection{Problem setting}
\label{sec:problem}
Given a set of $B$ image-report pairs, $\mathcal{D}=\{(\mathcal{I}_1, \mathcal{R}_1),\ldots, (\mathcal{I}_B, \mathcal{R}_B)\}$, we aim to train a multi-modal model to diagnose the presence of certain diseases in downstream datasets. At inference time, we can query the likelihood of a given disease category $c$, which is either seen or unseen during training, in the image $\mathcal{I} \in \mathbb{R}^{H\times W \times 3}$ by prompting the textual disease description $T_c$:
\begin{equation}
    \hat{p}, \hat{m} = \Phi_{\text{fusion}}(f_{\text{vision}}(\mathcal{I}), f_{\text{text}}(T_c)),
\end{equation}
where $f_\text{vision}$, $f_\text{text}$, and $\Phi_{\text{fusion}}$ refer to vision, language and fusion modules. Here, $\hat{p}$ is the predicted likelihood of the disease; and $\hat{m} \in \mathbb{R}^{H \times W}$ denotes a heatmap with high activation on pixels indicating disease's visual presence. This heatmap is used for zero-shot visual grounding.

\noindent \textbf{Reformulate VLP as a multi-label recognition.} As clinical reports pose linguistic challenges (e.g., negation expressions, dense clinical terminologies), contrastively matching with raw reports struggles to capture discriminative features. Given the availability of well-established medical entity tagging models, such as RadGraph~\cite{jain2021radgraph}, we reformulate VLP as a supervised multi-label recognition. Inspired by~\cite{wu2023medklip}, we adopt RadGraph~\cite{jain2021radgraph} to extract entities (\ie, diseases or any medical findings) in the reports: 
\begin{equation}
\label{eq:triplet}
    \phi(\mathcal{R}_i) = \{\text{entity}_t, \text{location}_t, \text{exist}_t\}, t \in [0, t_i],
\end{equation}
where $t_i$ refers to the number of medical entities in the report, and $\text{location}_t$ is the location (\eg lung, spine) where the $\text{entity}_t$ occurs in the image $\mathcal{I}_i$. 

Let $\mathcal{C}=\{c_1, \ldots, c_N\}$ and $L=\{l_1, \ldots, l_M\}$ denote the disease entities and locations extracted from all the report databases. 
Given an image $\mathcal{I}$, we reformulate VLP as a multi-label classification by directly predicting the presence of all diseases $\hat{p}_i \in \mathbb{R}^N$. Here, the ground-truth will be $Y_i=\{y_{i,1}, \ldots, y_{i,N}\}$, where $y_{i,j}=1$  if the entity $c_j \in \mathcal{C}$ is indicated by the positive presence, \ie $\text{exist}_j=1$, in the report $\mathcal{R}_i$ and 0 otherwise.  

\begin{figure}
    \centering
    \includegraphics[width=\linewidth]{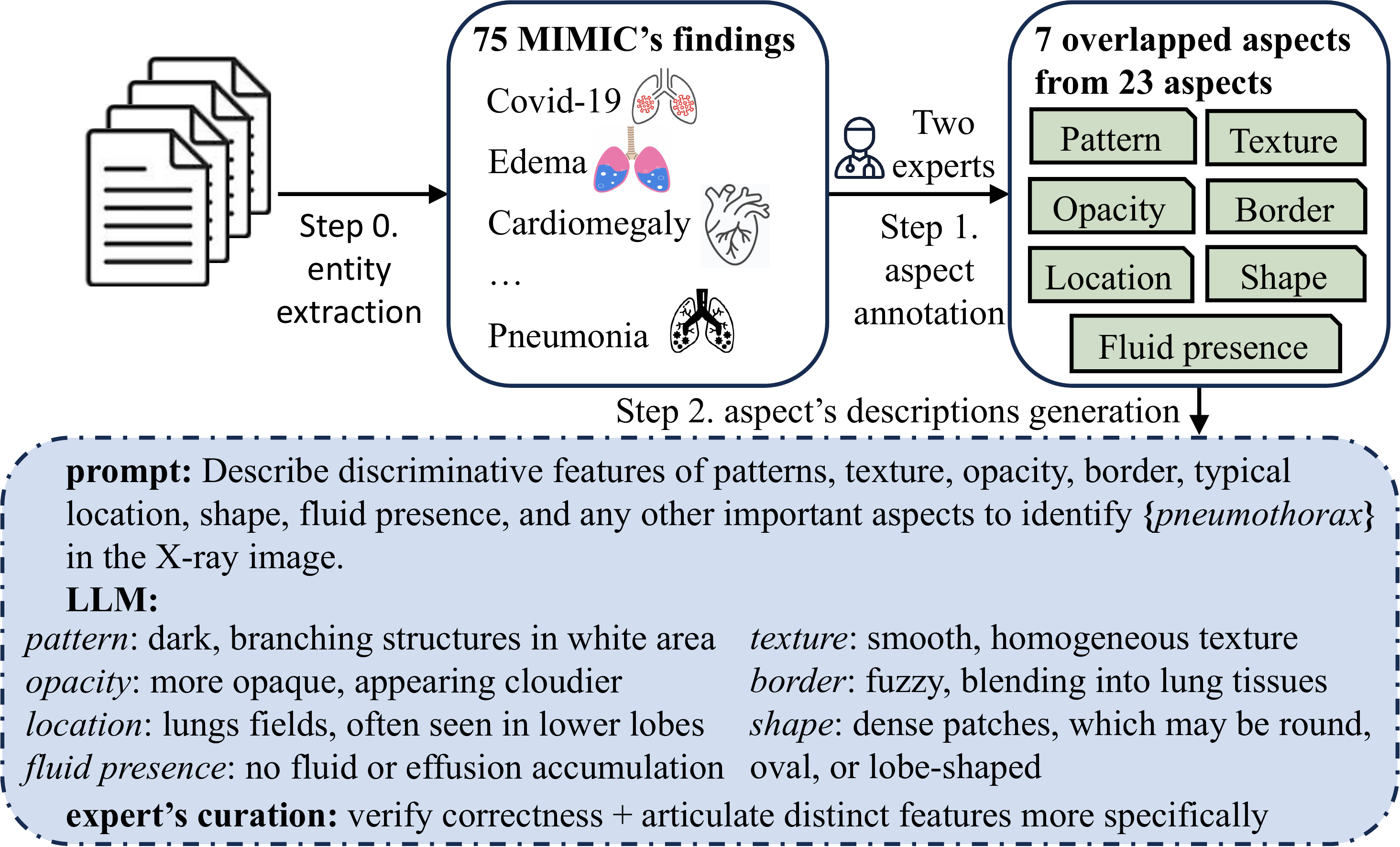}
    \caption{Pipeline to extract visual aspect's descriptions of diseases mentioned in the pre-training MIMIC dataset~\cite{johnson2019mimic}.}
    \label{fig:prompt}
    \vspace{-0.8em}
\end{figure}
\subsection{Multi-aspect vision-language pre-training for enhanced pathology detection} 
\label{sec:aspect}
\begin{figure*}[!h]
    \centering
    \includegraphics[width=\linewidth]{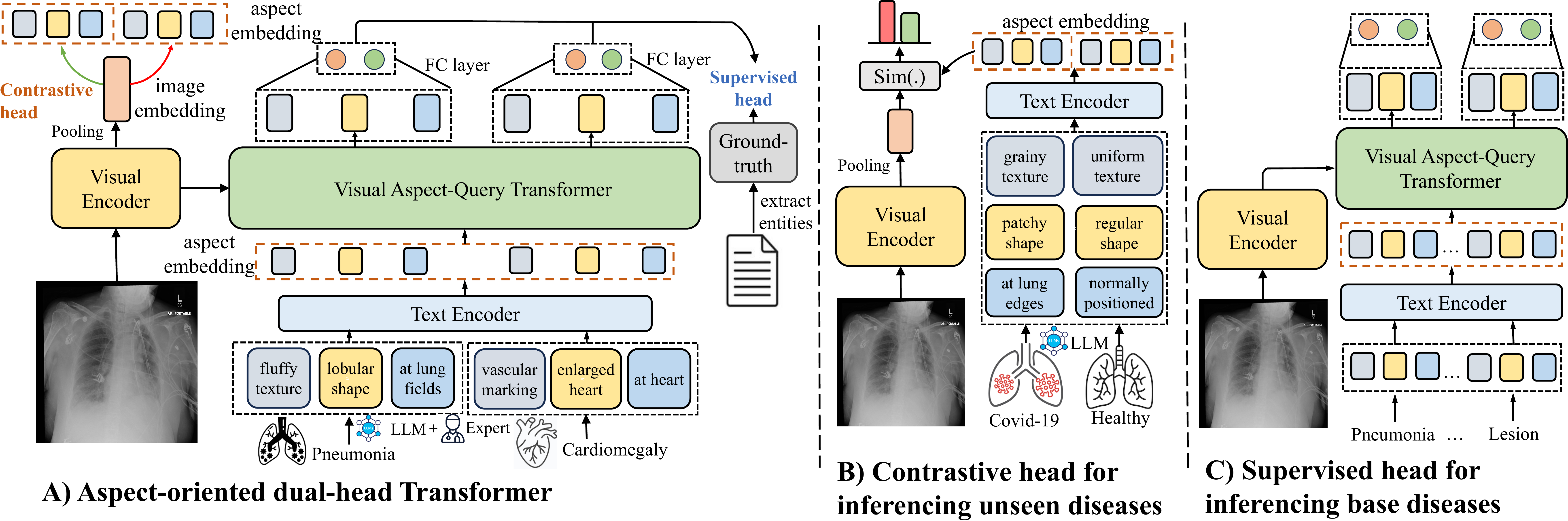}
    \caption{\textbf{Multi-aspect vision-language pre-training framework (MAVL)} decomposes diseases into a set of shared visual aspects using LLM and prior knowledge from two medical experts. An aspect-oriented dual-head Transformer (A) visually searches for the queried visual aspects in the image and maximizes detection accuracy of both unseen and seen diseases via two learning signals. The contrastive head (B) captures generalizable features and performs zero-shot classification of unseen diseases by comparing visual aspects between the target disease and the healthy category. The supervised head (C) captures discriminative features to classify fine-grained seen diseases.}
    \vspace{-1em}
    \label{fig:method}
\end{figure*}

To improve the image compatibility with the pathological representation, our paper proposes to dissect the disease description into a set of visual aspects by leveraging the prior knowledge from medical experts, and a large language model (LLM). For each disease entity $c_j$, our proposed framework decomposes into a set of $K$ visual aspects $S_j=\{s_{j1}, \ldots, s_{jK}\}$ by querying a knowledge base from medical professionals. We generate a description of each aspect entity $\text{Description}(s_{jk})$ using an LLM. To further differentiate diseases, we also add a fine-grained description of a disease $s_{j0}=\text{Description}(c_j)$, extracted from the UMLS knowledge base~\cite{bodenreider2004unified}. Our MAVL framework performs a fine-grained matching between an image $\mathcal{I}_i$ and every descriptive aspect $S_j$ of the target disease $c_j$ to detect its presence. An ablation study in Sec.~\ref{fig:method} analyzes which aspect yields higher performance gain.

This paper presents a semi-automatic pipeline to decompose and describe disease's visual aspects from the 75 medical findings extracted by RadGraph~\cite{jain2021radgraph}. Fig.~\ref{fig:prompt} depicts our proposed pipeline leveraging medical experts' knowledge base and the LLM. To reduce annotation bias, two medical experts, a board-certificated cardiologist and a radiologist, independently annotated the visual aspects of 75 medical findings. Two experts independently annotate 23 visual aspects in total (see Suppl.), among which 7 aspects are in consensus, as shown in Fig.~\ref{fig:prompt}. We also add an extra `other' aspect for the experts and the LLM to fill in any extra distinct features to recognize the target disease. In total, we have 8 visual aspects, plus a fine-grained disease description $s_{j0}$, which is extracted from the UMLS~\cite{bodenreider2004unified}

Secondly, we use GPT-4 to programmatically generate descriptions of the 8 annotated visual aspects. The pseudo-code is presented in the Supplementary. Finally, we consult two medical experts to curate the GPT-generated visual descriptions, which involves two steps. First, as LLMs can hallucinate contents, the experts correct the information. For example, GPT falsely describes `pneumothorax' as `more opaque, less transparent', which is then corrected as `less opaque, more transparent'. Second, we articulate the distinct features of diseases to disambiguate fine-grained categories. For instance, GPT-generated descriptions can be vague, e.g., `edema' has ``hazy and patchy textures". We add more unique characteristics of edema: ``fine, grainy, or mottled texture within the cloudy area, looking like small speckled spots". 



\subsection{Dual-head Transformer network.} 
\label{sec:transformer}
This section presents our proposed aspect-oriented dual-head Transformer network, which takes the decomposed visual aspects of each disease and searches for their presence in the image. Given the image $\mathcal{I}$, the visual encoder $f_{\text{vision}}$ extracts the visual features $\mathcal{V}$:
\begin{equation}
    \mathcal{V} = f_{\text{vision}}(\mathcal{I}) \in \mathbb{R}^{h \times w \times d},
\end{equation}
where $h$, $w$, and $d$ refer to the height, width and channel dimensions of the feature map. For a fair comparison with a recent work~\cite{wu2023medklip}, we adopt ResNet-50 for simplicity and computational efficiency. Yet, our framework is network-agnostic, and ViT backbones~\cite{dosovitskiy2020image} can be applied.

Given an entity $c_j$, and its $K$ visual aspects $S_j=\{s_{j1}, \ldots, s_{jK}\}$, the text encoder produces a set of $d$-dimension aspect embeddings $\mathbf{A}_j \in \mathbb{R}^{K \times d}$ for disease $c_j$:
\begin{equation}
    \mathbf{A}_j = \{f_{\text{text}}(\text{Description}(s_{jk})) \quad | \quad k \in [0, \ldots, K]\}.
\end{equation}
Capitalizing on the aspect representations, we develop a dual-head network with decoupled losses: a contrastive head for zero-shot recognition of unseen diseases, and a supervised head to maximize fine-grained classification of base diseases, as depicted in Fig.~\ref{fig:method}(A). 

\noindent \textbf{Contrastive head for zero-shot recognition of unseen diseases.} This head performs fine-grained (FG) contrastive learning by matching an image representation $\mathbf{F}$ with each aspect representation $\mathbf{a}_{jk}$ of all disease $c_j$. The contrastive head first pools the image representation $\mathbf{f}$ from the visual feature map $\mathbf{\mathcal{V}}$ using the attention pooling from CLIP~\cite{radford2021learning}:
\begin{equation}
    \mathbf{f} = \text{AttentionPool}(\mathcal{V}) \in \mathbb{R}^{d}.
\end{equation}
Given an image feature $\mathbf{f}_i$, we compute a FG contrastive loss $\mathcal{L}_{\text{cl}}$ with each $k$-th aspect $s_{jk}$ of every disease $c_j$:

\begin{equation}
    \mathcal{L}_{\text{cl}} = \sum_{k=1}^{K} \log \sum_{j \in P(i)}\frac{\exp{(\mathbf{f}^T_i\mathbf{a}_{jk}})}{\sum_{t}{\exp{(\mathbf{f}^T_i\mathbf{a}_{tk}})}}.
\end{equation}
Here, $P(i)=[c_j | y_{i,j}=1]$ denotes the set of diseases $c_j$ that are positively present in the report $\mathcal{R}_i$. During inferencing, our model programmatically extracts visual descriptions of new diseases using GPT-4, and compares its similarity with visual appearances of healthy versus pathological categories, as shown in Fig.~\ref{fig:method}(B). 

\textit{Discussion.} Diverging from previous report matching methods~\cite{boecking2022making,bannur2023learning,tiu2022expert}, ours matches with the textual representations of a common visual knowledge base (\eg, disease's shape, opacity levels). This allows the model to link attributes of new diseases with this knowledge base and captures extensible visual representations.  

\noindent \textbf{Supervised head for zero-shot recognition of base diseases.} 
Contrastive learning empirically struggles to extract discriminative features, leading to sub-optimal classification of fine-grained seen diseases. To optimize the fine-grained classification of a large set of base categories mentioned in the pre-training dataset, we develop supervised Transformer-based module $\Theta_{\text{ground}}$ to extract discriminative features by grounding the textual aspects in the image. 

To extract features for a disease entity $c_j$, we view its descriptive visual aspects $\mathbf{A}_j \in \mathbb{R}^{K \times d}$ as a disease query set $Q_j$. Note, we omit the sample index $i$ for simplicity. A Transformer-based fusion module $\Theta_{\text{ground}}$ grounds a query set $Q_j$ of textual visual aspects with image representation $\mathcal{V}$ using multiple Transformer Decoder layers~\cite{carion2020end}. The cross-attention views $Q_j$ as Query, and $\mathcal{V}$ as Key and Value. The vision-text grounded features $\mathcal{F}_j \in \mathbb{R}^{K \times d}$ for disease $c_j$ is formulated as
\begin{equation}
    \mathcal{F}_j, \tilde{m}_j  = \Theta_{\text{ground}}(Q_j, \mathcal{V}),
\end{equation}
where $\tilde{m}_j$ is the heatmap for visual grounding.  The feature maps are flattened into $\mathcal{F}_j \in \mathbb{R}^{Kd}$ and fed into a fully-connected layer parameterized by $W \in \mathbb{R}^{Kd \times 2}$ to predict the binary outcome of the disease $\hat{p}_j$. To detect all diseases, our fusion module takes a query set $Q=\{Q_1, \ldots, Q_N\}$ from all $N$ medical findings that have been mentioned in the pre-training report sets, extracts the grounded features, and classifies their presence $\hat{p} \in \mathbb{R}^{N \times 2}$.

By reformulating VLP as a multi-label classification (Sec.~\ref{sec:problem}), we directly train the visual aspect-query Transformer using supervision labels $Y$, thus improving zero-shot classification of seen diseases on downstream tasks. The supervised head is trained via a cross entropy loss $\mathcal{L}_{\text{sup}} = \text{CE}(\hat{p}, Y)$. Besides, the network predicts the location embedding $\tilde{e}_j$ of each disease from its features $\mathcal{F}_j$ using a simple projection layer. 
If the disease $c_j$ is positively present and appears at location $l_j$, we adopt the location contrastive loss $\mathcal{L}_{\text{loc}}$ from~\cite{wu2023medklip} to match the predicted location embedding $\tilde{e}_j$ 
with the positive location embedding $e_j = f_{\text{text}}(\text{``It is located at } \{l_j\}\text{"})$.  
 The network is trained end-to-end with the combined loss objectives:
\begin{equation}
    \mathcal{L} = \alpha \mathcal{L}_{\text{cl}} + \beta \mathcal{L}_{\text{sup}} + \gamma \mathcal{L}_{\text{loc}}.
\end{equation}

During inferencing, if the target category belongs to the set of $N$ base categories $\mathcal{C}$, the model will use a supervised head for inferencing, as depicted in Fig.~\ref{fig:method}(C).


\textit{Discussion}. While MedKLIP~\cite{wu2023medklip} also converts VLP into a supervised learning framework, we differ in two folds. \textit{First}, they represent each entity $c_j$ as a generic medical description with complex clinical terminologies (refer to Fig.~\ref{fig:illustrate}). In contrast, our framework dissects the disease into a set of descriptive aspects, thus improving the image compatibility with the pathological representation. Importantly, we enable effective recognition of new diseases by translating them into elemental aspects, shared with base diseases. \textit{Second}, MedKLIP~\cite{wu2023medklip} excludes the contrastive loss between image and text representations. While achieving high performance on base categories, their supervised model falters in unseen categories. In contrast, we decouple learning signals and introduce a dual-head network, tailored to process both known and unknown diseases.

%% file: sec/4_result.tex
\begin{table*}[tp]
\centering
\caption{Comparisons with SOTA image-text pre-training models under zero-shot classification for base diseases that have been seen during pre-training across 5 datasets.}
\label{tab:seen-cls}
\resizebox{1.0\textwidth}{!}{
\begin{tabular}{c|ccc|ccc|ccc|ccc|ccc}
\toprule
\textbf{Dataset} & \multicolumn{3}{c}{\textbf{CheXpert}~\cite{irvin2019chexpert}} & \multicolumn{3}{c}{\textbf{ChestXray-14}~\cite{wang2017chestx}}  & \multicolumn{3}{c}{\textbf{PadChest-seen}~\cite{bustos2020padchest}}  & \multicolumn{3}{c}{\textbf{RSNA Pneumonia}~\cite{shih2019augmenting}} & \multicolumn{3}{c}{\textbf{SIIM-ACR}~\cite{siim2019}} \\
\cmidrule(lr){2-4} \cmidrule(lr){5-7} \cmidrule(lr){8-10} \cmidrule(lr){11-13} \cmidrule(lr){14-16}
\textbf{Method} & AUC $\uparrow$ & F1 $\uparrow$ & ACC $\uparrow$ & AUC $\uparrow$ & F1 $\uparrow$ & ACC $\uparrow$ & AUC $\uparrow$ & F1 $\uparrow$ & ACC $\uparrow$ & AUC $\uparrow$ & F1 $\uparrow$ & ACC $\uparrow$ & AUC $\uparrow$ & F1 $\uparrow$ & ACC $\uparrow$  \\
\midrule
ConVIRT~\cite{zhang2022contrastive} & 52.10 & 35.61 & 57.43 & 53.15 & 12.38 & 57.88 & 63.72 & 14.56 & 73.47 & 79.21 & 55.67 & 75.08 & 64.25 & 42.87 & 53.42 \\
GLoRIA~\cite{huang2021gloria} & 54.84 & 37.86 & 60.70 & 55.92 & 14.20 & 59.47 & 64.09 & 14.83 & 73.86 &  70.37 & 48.19 & 70.54 & 54.71 & 40.39 & 47.15 \\
BioViL~\cite{boecking2022making} & 60.01 & 42.10 & 66.13 & 57.82 & 15.64 & 61.33  & 60.35 & 10.63 & 70.48 & 84.12 & 54.59 & 74.43 & 70.28 & 46.45 & 68.22 \\
BioViL-T~\cite{bannur2023learning} & 70.93 & 47.21 & 69.96 & 60.43 & 17.29 & 62.12  & 65.78 & 15.37 & 77.52  & 86.03 & 62.56  & 80.04  & 75.56 & 60.18 & 73.72 \\
CheXzero~\cite{tiu2022expert} & 87.90 & 61.90 & 81.17 & 66.99 & 21.99 & 65.38  & 73.24 & 19.53  & 83.49 & 85.13 & 61.49 & 78.34 & 84.60 & 65.97 & 77.34 \\
MedKLIP~\cite{wu2023medklip} & 87.97 & 63.67 &  84.32  & 72.33 & 24.18 & 79.40  & 77.87 & 26.63 & 92.44  & 86.57 & 63.28 & 79.97 & 89.79 & 72.73 & 83.99 \\\midrule 
Ours & \textbf{90.13} & \textbf{65.47} & \textbf{86.44} & \textbf{73.57} & \textbf{26.25} & \textbf{82.77} & \textbf{78.79} & \textbf{28.48} & \textbf{92.56} & 
\textbf{86.91}& \textbf{63.41} & \textbf{82.42} & \textbf{92.04} & \textbf{77.95} & \textbf{87.14}  \\
\bottomrule
\end{tabular}
}
\end{table*}

\section{Experimental Setting}
\subsection{Pre-training dataset}
\textbf{MIMIC-CXR v2}~\cite{johnson2019mimic} consists of more than 227k studies comprising paired image-report data derived from 65,379 distinct patients who underwent scanning procedures. Each individual study may contain either one or two images, representing different scan perspectives, resulting in a cumulative dataset of 377,110 images.

\subsection{Datasets for downstream tasks}
\textbf{ChestX-ray14}~\cite{wang2017chestx} comprises 112,120 frontal-view X-ray images from 30,805 individual patients, collected by NIH (National Institutes of Health) from 1992 to 2015. It includes labels for 14 prevalent diseases. We partition it into 0.8/0.1/0.1 for train/valid/test.

\noindent\textbf{CheXpert}~\cite{irvin2019chexpert} includes 224,316 chest X-ray images from 65,240 patients collected at Stanford Hospital. The official validation set comprises 200 chest radiographic studies annotated by three board-certified radiologists, while the official test set contains 500 studies annotated by five board-certified radiologists. Following the evaluation procedure from previous works~\cite{tiu2022expert,irvin2019chexpert}, we evaluate five observations in the official test set for the competition tasks. 

\noindent\textbf{PadChest} comprises 160,868 chest X-ray images obtained from 67K patients reported at Hospital San Juan Hospital (Spain). They are annotated with 150+ distinct radiographic findings, including both unseen and seen classes during pre-training. 
We denote the 14 seen diseases as \textit{PadChest-seen}, and unseen ones as \textit{PadChest-unseen}. Among novel classes, we also evaluate rare diseases by selecting the top 20 diseases with the lowest number of samples, and the pathology recorded in the National Organization of Rare Disease (NORD) database\footnote{\href{https://rarediseases.org/rare-diseases/}{https://rarediseases.org/rare-diseases/}}, leading to 39 rare diseases in total. Approximately 27\% of the data (39,053 examples) are annotated by radiologists, while the remainder are generated by a recurrent neural network. We only report the results on expert-annotated test samples.

\noindent\textbf{RSNA Pneumonia}~\cite{shih2019augmenting} includes over 260,000 frontal-view chest X-rays with the annotated pneumonia masks collected by the Radiological Society of North America (RSNA). The dataset can be used for both pneumonia segmentation and classification tasks~\cite{huang2021gloria,wu2023medklip,boecking2022making}. We split it into 0.6/0.2/0.2 for train/valid/test.

\noindent\textbf{SIIM-ACR Pneumothorax}~\cite{siim2019} comprises over 12,000 frontal-view chest X-rays with pneumothorax masks, collected by the Society for Imaging Informatics in Medicine and the American College of Radiology (SIIM-ACR). We split it into 0.6/0.2/0.2 for train/valid/test.

\noindent\textbf{COVIDx CXR-2}~\cite{pavlova2022covid} and \textbf{COVID Rural}~\cite{desai2020chest} provide benchmark dataset for Covid-19 detection. COVIDx CXR-2~\cite{pavlova2022covid} consists of 29,986 images from 16,648 COVID-19 patients, labeled for classification. Following~\cite{wu2023medklip}, we split it into 0.7/0.2/0.1 for train/valid/test set. COVID Rural~\cite{desai2020chest} consists of over 200 chest X-ray with annotated segmentation masks for COVID-19, which is used for segmentation evaluation. We partition it into 0.6/0.2/0.2 for train/valid/test set.

\subsection{Evaluation metrics} Following previous works~\cite{wu2023medklip,bannur2023learning,boecking2022making}, we adopt standard metrics for classification, including AUC scores, F1 scores, and accuracy, and those for segmentation, including Dice score, IoU scores, and pixel-wise accuracy. The metrics refer to the macro average on all the diseases present in the target dataset. All metrics are reported in percentage (\%).

\subsection{Implementation details} In \textit{pre-training}, the text encoders are initialized with the weights of ClinicalBERT~\cite{alsentzer2019publicly} and frozen. Only the visual encoder and the Transformer fusion module are trained during pre-training. All models are trained on 4 A100 GPUs. In \textit{fine-tuning}, following previous works~\cite{wu2023medklip,bannur2023learning,boecking2022making}, we use ResNet50~\cite{he2016deep} for classification, and ResUNet~\cite{diakogiannis2020resunet} for segmentation, which will be initialized with our pre-trained visual encoder. More details about hyper-parameter analysis can be found in the supplementary material.

\begin{table*}[tp]
\hspace{-0.1em} 
\parbox[t]{.63\textwidth}{
\footnotesize
\centering
\setlength{\tabcolsep}{8pt}
\begin{threeparttable}
\begin{tabular}{c|ccc|cc|cc}
\toprule
\textbf{Dataset}  & \multicolumn{3}{c}{\textbf{Covid-19 CXR-2}} & \multicolumn{2}{c}{\textbf{PadChest-unseen}} & \multicolumn{2}{c}{\textbf{PadChest-rare}} \\
\cmidrule(lr){2-4} \cmidrule(lr){5-6} \cmidrule(lr){7-8}
\textbf{Method} & AUC $\uparrow$ & F1 $\uparrow$ & ACC $\uparrow$  & AUC $\uparrow$ & ACC $\uparrow$ & AUC $\uparrow$ & ACC $\uparrow$ \\
\midrule
ConVIRT~\cite{zhang2022contrastive} & 62.78 & 71.23 & 63.84 & 51.17 & 61.51 & 50.37 & 60.17 \\
GLoRIA~\cite{huang2021gloria} & 64.52 & 70.78 & 60.21 & 49.96 & 60.95 & 48.25 & 58.49\\
BioViL~\cite{boecking2022making} & 61.40 & 70.92 & 58.20 & 57.95 & 62.50 & 52.82 & 60.60 \\
BioViL-T~\cite{bannur2023learning} & 62.43 & 69.64 & 57.65 & 58.94 & 68.56 & 57.44 & 65.38  \\
CheXzero~\cite{tiu2022expert} & 73.13 & 76.13 & 71.45 & 66.70 & 81.19 & 65.08 & 81.17 \\
MedKLIP~\cite{wu2023medklip} & 76.28 & 76.54 & 71.96 & 60.31 & 76.69 & 59.75 & 77.84 \\\midrule  
MAVL - Sup & \ul{78.74} & \ul{80.21} & \ul{75.08} & \ul{67.84} & \ul{80.22} & \ul{67.20} & \ul{82.94} \\
MAVL - Con & \textbf{83.86} & \textbf{81.73} & \textbf{78.07} & \textbf{70.42} & \textbf{84.00} & \textbf{70.06} & \textbf{84.64} \\\bottomrule
\end{tabular}
\end{threeparttable}
}
\hspace{\fill}
\parbox[t]{.31\textwidth}{
\caption{Zero-shot classification results of unseen classes on the two datasets. Results on PadChest's rare diseases are reported. PadChest (from Spain) has a large distribution shift from the pre-training MIMIC data (from US), 
which is challenging for zero-shot classification. Following a prior work~\cite{tiu2022expert}, we only consider AUC scores and accuracy. `Sup' and 'Con' denote supervised and contrastive heads.}
\label{tab:unseen-cls}
}
\vspace{-0.6cm}
\end{table*}

\section{Experimental results and discussion}
\label{sec:result}
This section presents the experimental results for zero-shot and fine-tuning setting. In the zero-shot case (Sec.~\ref{sec:zero}), we benchmark our method against SOTA models for seen and unseen diseases on classification and segmentation tasks. In the fine-tuning case (Sec.~\ref{sec:ft}), we evaluate the model's transferability on classification and segmentation tasks. 
\subsection{Zero-shot evaluation}
\label{sec:zero}
\textbf{Classification for seen diseases.}
Table~\ref{tab:seen-cls} presents the zero-shot classification benchmark of \textit{seen} diseases. Supervised methods, including MedKLIP and ours surpass other contrastive methods for base disease classification, showing that contrastive learning is non-optimal to capture discriminative features for fine-grained base disease classification. Compared to MedKLIP, our model consistently improves F1 scores 
by up to 8.56\% across 5 datasets, collected from various demographics, e.g., US population for CheXpert~\cite{irvin2019chexpert} and RSNA~\cite{shih2019augmenting}, and Spanish for PadChest~\cite{bustos2020padchest}. By providing detailed visual aspects, our method excels at disambiguating fine-grained diseases, offering a robust foundational model against different imaging distributions.

\noindent \textbf{Classification for unseen diseases.} Table~\ref{tab:unseen-cls} presents the zero-shot classification benchmark of unseen diseases. We include the performance of both supervised and contrastive heads in our model for comparisons. We note three observations. \textit{First}, both our heads consistently outperform previous methods on both datasets. Our supervised head drastically improves the AUC of a contrastive CheXzero, and a supervised MedKLIP by up to 7.67\% and 12.49\%. By decomposing the unstructured disease description into a common set of structured elemental aspects, our model effectively links new disease aspects with base diseases, improving feature representations of unseen diseases. 

\textit{Second}, contrastive models are generally more effective for novel disease recognition than the supervised counterparts. While the supervised MedKLIP struggles to classify many unseen diseases in PadChest, our contrastive head improves the AUC score of MedKLIP and CheXzero by 16.76\% and 5.58\%, respectively. This shows the effectiveness of our dual-head Transformer design, tailored to process both known and unknown diseases. \textit{Third}, both of our heads surpass previous methods in classifying rare diseases. Our contrastive head significantly improves the AUC scores of MedKLIP and CheXzero by 17.26\% and 7.65\%. Leveraging prior knowledge about pathological visual profiles boosts the classification of the long-tailed distribution. 

\begin{figure}[!h]
    \centering
    \includegraphics[width=\linewidth]{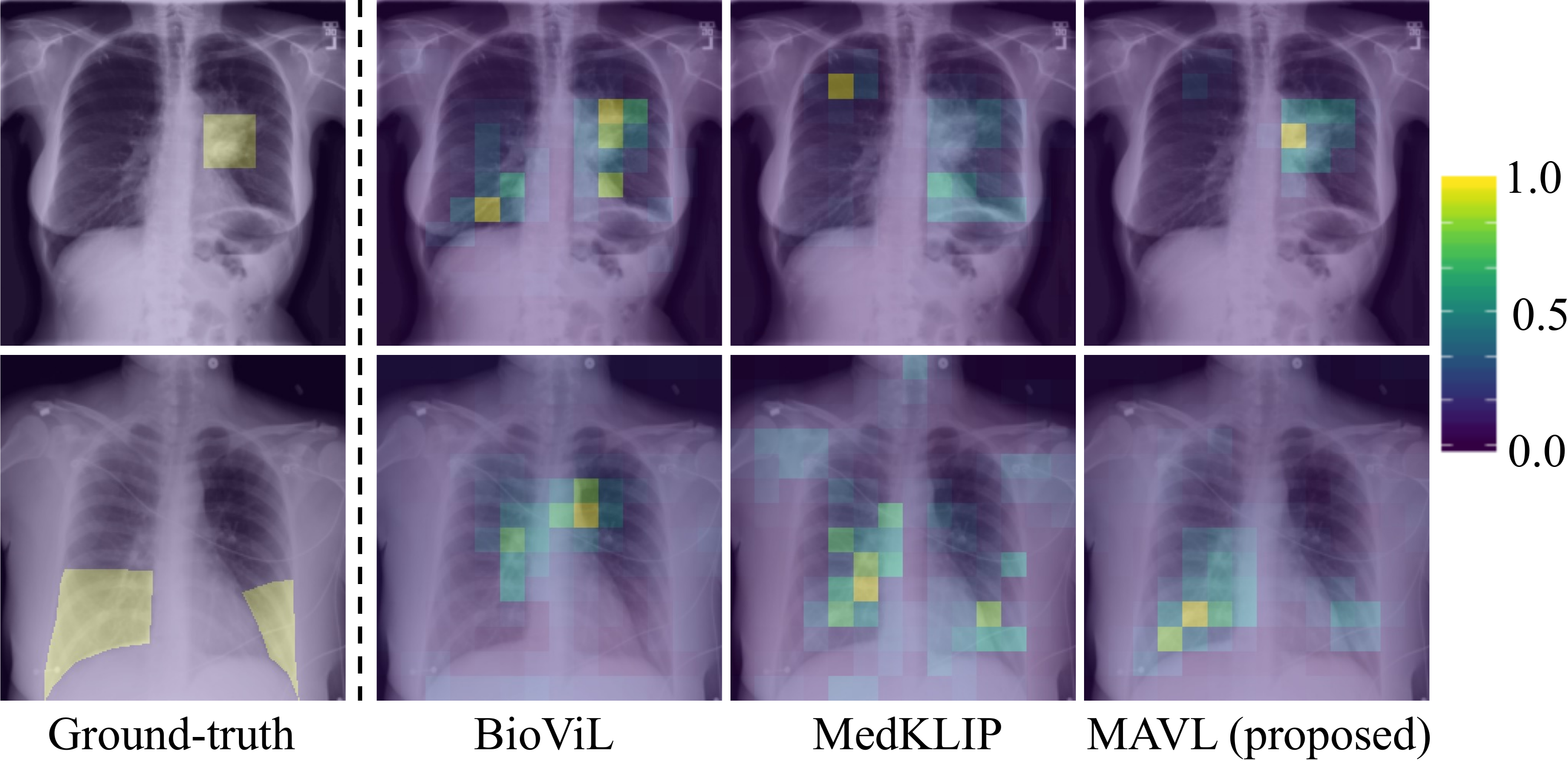}
    \caption{Visual grounding prediction scores of RSNA Pneumonia (Top) and Covid-19 (Bottom).}
    \label{fig:vis}
    \vspace{-0.7em}
\end{figure}

\noindent \textbf{Visual grounding.} In addition to the diagnosis, explainability is equally important in the medical domain. To this end, our model offers explainability by grounding the abnormal regions in the image. Table~\ref{tab:seg} presents visual grounding results of different models for both \textit{seen} and \textit{unseen} diseases:  Pneumonia on RNSA dataset~\cite{shih2019augmenting} and Covid-19 on Covid-19 Rural dataset~\cite{desai2020chest}. 
Our model consistently surpasses previous methods on visual grounding. By dissecting disease entities into descriptive aspects, our model exhibits enhanced effectiveness in localizing abnormalities in images. 
In particular, we improve the Dice score of MedKLIP on a novel disease, Covid-19 by 5.34\%. Decomposing the disease into a set of common visual aspects as in our MAVL improves the generalization of unseen diseases. Fig.~\ref{fig:vis} shows the visual grounding results of different methods on Pneumonia and Covid-19. Our model precisely localizes the abnormality, yielding lower false positive detection compared to other methods. 

\begin{table}[!h]
\centering
\caption{Comparison with SOTA image-text pre-training models on zero-shot region grounding tasks on Pneumonia and Covid-19.}
\label{tab:seg}
\resizebox{0.47\textwidth}{!}{
\begin{tabular}{c|ccc|ccc}
\toprule
\textbf{Dataset}  & \multicolumn{3}{c}{\textbf{RSNA Pneumonia}~\cite{shih2019augmenting}} & \multicolumn{3}{c}{\textbf{Covid-19 Rural}~\cite{desai2020chest}} \\
\cmidrule(lr){2-4} \cmidrule(lr){5-7}
\textbf{Method} & IoU $\uparrow$ & Dice $\uparrow$ & ACC $\uparrow$  & IoU $\uparrow$ & Dice $\uparrow$ & ACC $\uparrow$ \\
\midrule
GLoRIA~\cite{huang2021gloria} & 21.82 & 34.68 & 75.14 & 8.18 & 12.49 & 66.73 \\
BioViL~\cite{boecking2022making} &  30.29 & 43.86 & 82.15 & 11.52 &  15.77 & 70.48\\
MedKLIP~\cite{wu2023medklip} & 34.41 & 49.23 & 86.90 & 20.88 & 32.38 & 76.23\\\midrule
MAVL (ours) & \textbf{34.72} & \textbf{50.04} & \textbf{88.53} &  \textbf{21.97} & \textbf{34.11} & \textbf{84.29} \\\bottomrule
\end{tabular}
}
\vspace{-0.3em}
\end{table}

\begin{table*}[tp]
\centering
\caption{Results of different VLP models under fine-tuning classification and segmentation with different data portions. AUC scores and Dice scores are respectively reported for the two tasks.}
\label{tab:ft}
\resizebox{1.0\textwidth}{!}{
\begin{tabular}{c|ccc|ccc|ccc|ccc||ccc|ccc}
\toprule
Task & \multicolumn{12}{c||}{Classification} & \multicolumn{6}{c}{Segmentation} \\ \cmidrule(lr){2-13} \cmidrule(lr){14-19}
Dataset & \multicolumn{3}{c}{Pneumonia~\cite{shih2019augmenting}} & \multicolumn{3}{c}{Pneumothorax~\cite{siim2019}} & \multicolumn{3}{c}{Covid-19 CXR-2~\cite{pavlova2022covid}} & \multicolumn{3}{c||}{ChestXray-14~\cite{wang2017chestx}} & \multicolumn{3}{c}{Pneumonia~\cite{shih2019augmenting}} & \multicolumn{3}{c}{Covid-19 Rural~\cite{desai2020chest}} \\ \cmidrule(lr){2-4} \cmidrule(lr){5-7} \cmidrule(lr){8-10} \cmidrule(lr){11-13} \cmidrule(lr){14-16} \cmidrule(lr){17-19}
Data portion & 1\% & 10\% & 100\% &  1\% & 10\% & 100\% &  1\% & 10\% & 100\% &  1\% & 10\% & 100\%  &  1\% & 10\% & 100\%  &  1\% & 10\% & 100\%\\
\midrule
Scratch & 68.94 & 83.31 & 87.12 & 53.11 & 76.18 & 87.48 & 85.11 & 93.65  & 98.86 & 45.88 & 56.27 & 67.03 & 45.29 &58.42 & 69.75 & 14.09 & 25.97 & 37.83 \\
ConVIRT~\cite{zhang2022contrastive} & 78.86 & 85.42 & 87.64 & 72.39 & 80.41 & 91.67 & 90.30 & 97.74 & 99.70 & 57.23 & 72.53 & 79.13 & 56.48 & 63.94 & 71.87 & 16.97 & 30.79 & 42.71 \\
GLoRIA~\cite{huang2021gloria} & 79.13 & 85.59 & 87.83 & 75.85 & 86.20 & 91.89 & 92.74 & 97.18 & 99.54 & 58.94 & 72.87 & 79.92 & 58.13 & 67.71 & 72.06 & 16.12 & 31.20 & 43.85 \\
BioViL~\cite{boecking2022making} & 80.27 & 86.04 & 88.29 & 70.29 & 79.45 & 88.05& 92.39& 98.39 & 99.68 & 60.83 & 72.94 & 80.16 & 60.25 & 68.72 & 72.18 & 17.65 & 37.75 & 47.34 \\
MedKLIP~\cite{wu2023medklip} & 82.11 & 87.14 & 88.58 & 85.24 & 89.91 & 93.02 & 95.58 & 98.77 & 99.77 & 62.09 & 74.02 & 80.90 & 62.36 & 70.24 & 73.88 & 18.58 & 39.28 & 48.65 \\
\bottomrule
MAVL (ours) & \textbf{86.09} & \textbf{87.90} & \textbf{88.94} & \textbf{91.53} & \textbf{93.00} & \textbf{94.48}  & \textbf{97.18} & \textbf{99.15} & \textbf{99.90} & \textbf{68.65} & \textbf{80.50} & \textbf{86.22} & \textbf{71.95} & \textbf{73.51} & \textbf{76.97} & \textbf{24.51} & \textbf{40.71} & \textbf{50.25}\\\bottomrule
\end{tabular}
}
\end{table*}

\subsection{Fine-tuning evaluation}
\label{sec:ft}
This section provides fine-tuning evaluations by using the pre-trained models of different VLP methods to initialize weights and fine-tune downstream datasets.
Table~\ref{tab:ft} presents fine-tuning results on 4 datasets with 1\%/10\%/100\% data portion, which is consistent with previous works~\cite{wu2023medklip,boecking2022making}. Our model yields consistent improvements over previous works across all settings, especially under 1\% data portion for both classification and segmentation. Notably, the proposed MAVL significantly outperforms the second-best method, MedKLIP, by 7.38\% and 15.38\% on SIIM Pneumothorax classification and RSNA-Pneumonia segmentation when fine-tuning on 1\% data. On a fine-grained ChestXray-14 with 14 lung diseases, our MAVL offers drastic performance gain even when fine-tuning with 100\% data. Injecting prior knowledge about the disease's visual aspects during pre-training improves the model's transferability under low-shot learning, especially on fine-grained disease recognition.

\subsection{Ablation study}
\label{sec:abl}
\begin{figure}[!h]
    \centering
    \includegraphics[width=\linewidth]{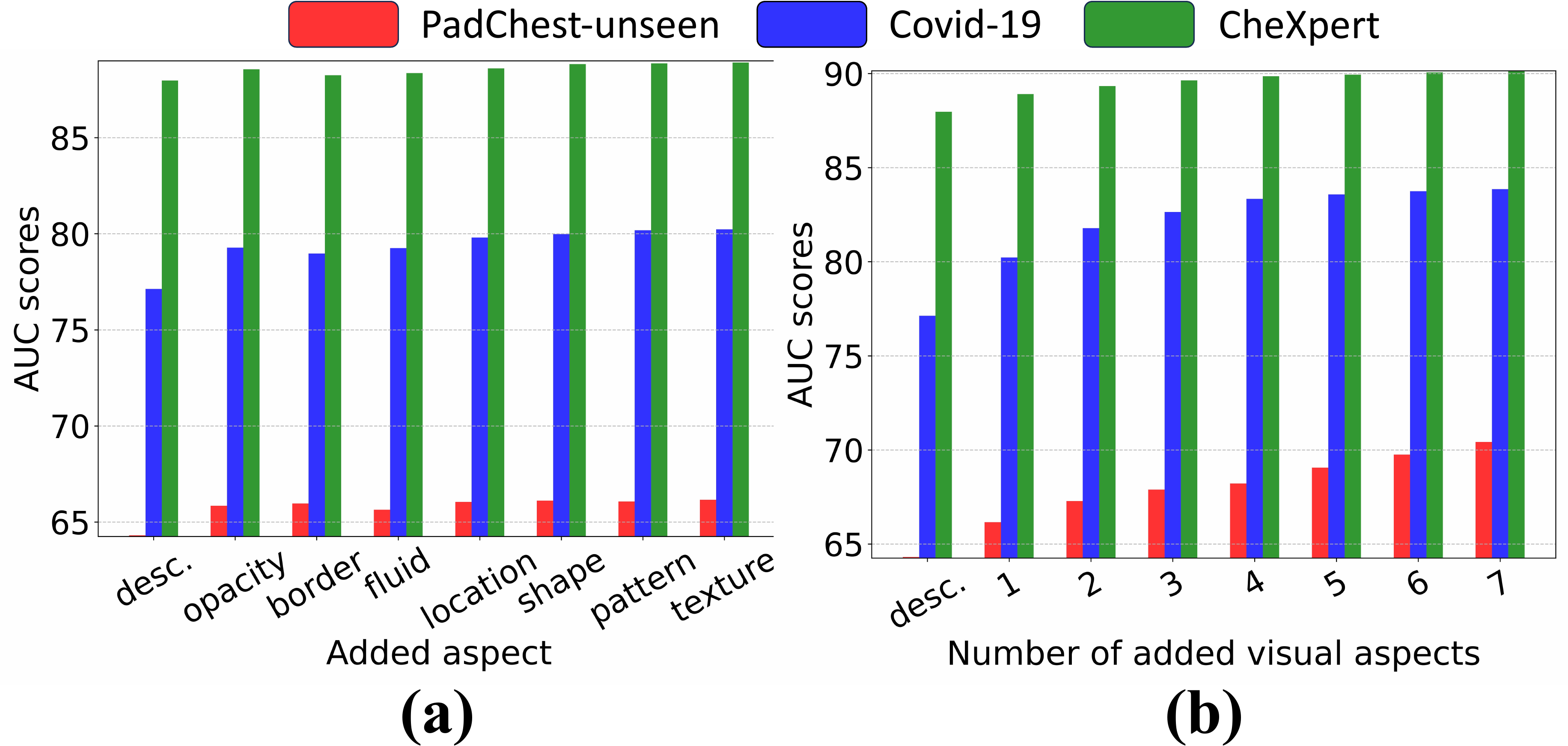}
    \caption{AUC scores of our method when (a) adding a single aspect, besides a disease clinical description and (b) when gradually adding higher numbers of aspects by their increasing importance order. The scores of using clinical description (desc.) of diseases are included as the baseline.}
    \label{fig:abl-aspect}
    \vspace{-1.0em}
\end{figure}

\textbf{Effectiveness of multi-aspect decomposition for VLP.} 
This section answers two questions: (i) whether adding more concepts improves zero-shot performance, and (ii) which concept matching yields higher performance gain. Fig.~\ref{fig:abl-aspect}(a) presents the performance when adding a single aspect. 
Matching with any single aspect boosts the accuracy of matching with a disease's definition as in~\cite{wu2023medklip}. Notably, the AUC score gain on unseen Covid CXR-2 is 4.02\% when using the aspect `texture'. 
More distinct aspects such as texture, patterns, and shapes yield higher  gains. 
Fig.~\ref{fig:abl-aspect}(b) shows the AUC scores when adding increasing numbers of visual aspects. Matching with higher numbers of aspects improves the zero-shot classification. 
    


\noindent \textbf{Effectiveness of a dual-head design.} This section analyzes the effectiveness of our dual-head network, which decouples contrastive and supervised learning. 
We implement a single-head model using only either a contrastive branch as in~\cite{tiu2022expert,bannur2023learning} or a supervised branch as in~\cite{wu2023medklip} 
\textit{First}, Table~\ref{tab:abl-decou} shows that using a single-head network design as in current literature compromises the performance of either seen and unseen classes. Capitalizing both branches in our dual-head design offers the highest F1 scores of 81.73\% and 26.25\% respectively on unseen Covid-19 and seen ChestXray-14. \textit{Second}, adding supervised learning in our Dual-Con boosts the AUC scores of the single contrastive head by 14.63\% on ChestXray-14. Similarly, adding CL in our Dual-Sup improves single supervised network. Integrating both learning signals as in our dual-head design complements the learning 
on both seen and unseen diseases. 


\begin{table}[!h]
\centering
\caption{Zero-shot classification results between a single-head (Single) network, and our dual-head network. Con and Sup denote contrastive and supervised heads of our dual-head network. 
}
\label{tab:abl-decou}
\resizebox{0.475\textwidth}{!}{
\begin{tabular}{cc|ccc|ccc}
\toprule
\multicolumn{2}{c}{Dataset} & \multicolumn{3}{c}{Covid-19 CXR-2~\cite{desai2020chest}} & \multicolumn{3}{c}{ChestXray-14~\cite{wang2017chestx}} \\
\cmidrule(lr){1-2}\cmidrule(lr){3-5} \cmidrule(lr){6-8}
Method & Loss & AUC $\uparrow$ & F1 $\uparrow$ & ACC $\uparrow$  & AUC $\uparrow$ & F1 $\uparrow$ & ACC $\uparrow$ \\
\midrule
Single - Con & CL &  81.65 & 80.75 & 76.98 & 62.08 & 18.27 & 62.53 \\
Dual - Con & CL/CE  & \textbf{83.86} & \textbf{81.73} & \textbf{78.07} & 71.16 & 25.58 & 75.81\\\midrule
Single - Sup & CE & 77.85 & 78.85 & 76.85 & 72.71 &  25.06 & 79.09 
\\
Dual - Sup & CL/CE & 78.74 & 80.21 & 75.08 & \textbf{73.57} & \textbf{26.25} & \textbf{82.77}\\\bottomrule
\end{tabular}
}
\end{table}
\vspace{-1.0em}

%% file: sec/5_conclusion.tex
\section{Conclusion}
This study introduces a Multi-Aspect Vision-Language Pre-training (MAVL) framework, designed to intricately dissect pathological entities into distinct, visually descriptive aspects, thereby significantly enhancing pathology detection. Our MAVL framework incorporates a dual-head, aspect-oriented Transformer, which decouples learning signals to optimize performance across both seen and unseen diseases. This approach is particularly effective for novel diseases, achieving a remarkable 16.76\% performance gain over existing methods. Moreover, MAVL's aspect-based matching leads to an impressive 17.26\% AUC improvement for rare unseen diseases.

\noindent \textbf{Acknowledgement.} This research was partially funded by the Hospital Research Foundation Group and supported by Dr. Perperidis from Adelaide's Women's and Children's Health Network to raise necessary funding.